\documentclass{article}
\usepackage{spconf,amsmath,graphicx}

\title{Ego-based Entropy Measures for Structural Representations on Graphs}
\name{George Dasoulas$^{\star\dagger}$, Giannis Nikolentzos$^{\star}$, Kevin Scaman$^{\dagger}$, Aladin Virmaux$^{\dagger}$ and Michalis Vazirgiannis$^{\star}$}
\address{$^{\star}$ {\'E}cole Polytechnique, Paris, France \\ $^{\dagger}$ Noah's Ark Lab, Huawei Technologies France}

\usepackage[ruled,vlined]{algorithm2e}
\usepackage{multirow}
\usepackage{subcaption}
\usepackage{hyperref}
\usepackage{url}
\usepackage{graphicx,wrapfig}  
\usepackage{nccmath}
\usepackage{amsmath}

\newcommand{\MN}{VNEstruct\xspace}

\makeatletter
\newcommand*{\addFileDependency}[1]{
  \typeout{(#1)}
  \@addtofilelist{#1}
  \IfFileExists{#1}{}{\typeout{No file #1.}}
}
\makeatother

\usepackage{amsmath,amsfonts,bm}

\def\Figref#1{Figure~\ref{#1}}

\def\eqref#1{equation~\ref{#1}}
\def\Eqref#1{Equation~\ref{#1}}

\def\1{\bm{1}}

\DeclareMathAlphabet{\mathsfit}{\encodingdefault}{\sfdefault}{m}{sl}
\SetMathAlphabet{\mathsfit}{bold}{\encodingdefault}{\sfdefault}{bx}{n}

\DeclareMathOperator*{\argmin}{arg\,min}

\DeclareMathOperator{\Tr}{Tr}

\begin{document}
%
\maketitle
\begin{abstract}
Machine learning on graph-structured data has attracted high research interest due to the emergence of Graph Neural Networks (GNNs). Most of the proposed GNNs are based on the node homophily, i.e neighboring nodes share similar characteristics. However, in many complex networks, nodes that lie to distant parts of the graph share structurally equivalent characteristics and exhibit similar roles (e.g chemical properties of distant atoms in a molecule, type of social network users). A growing literature proposed representations that identify structurally equivalent nodes. However, most of the existing methods require high time and space complexity. In this paper, we propose \MN, a simple approach, based on entropy measures of the neighborhood's topology, for generating low-dimensional structural representations, that is time-efficient and robust to graph perturbations. Empirically, we observe that \MN exhibits robustness on structural role identification tasks. Moreover, \MN can achieve state-of-the-art performance on graph classification, without incorporating the graph structure information in the optimization, in contrast to GNN competitors.

\end{abstract}
\begin{keywords}
Structural node representations, graph classification, graph neural networks
\end{keywords}
\section{Introduction}

The amount of data that can be represented as graphs has increased significantly in recent years. Graph representations are ubiquitous in several fields such as in biology, chemistry~\cite{pmlr-v70-gilmer17a}, and social networks~\cite{hamilton}.
Many applications require performing machine learning tasks on graph-structured data, such as graph classification~\cite{xu2018powerful}, semi-supervised node classification~\cite{Kipf:2016tc} and link prediction~\cite{kipf2016variational}.

The past few years have witnessed great activity in the field of learning on graphs. Graph Neural Networks (GNNs) have emerged as a successful approach on learning node-level and graph-level representations. 
Recently, the interest on this field has focused on the expressiveness of GNNs~\cite{xu2018powerful,clip} and how these models can be deep enough to extract long-range information from distant nodes in the graph~\cite{deeperinsight,pairnorm,khop}. So far, most of the models are designed so that they preserve the proximity between nodes, i.e., nodes that are close to each other in the graph obtain similar representations~\cite{hamilton, perozzi2014deepwalk}. However, some tasks require assigning similar representations to nodes that can be distant in the graph, but structurally equivalent. For example, in chemistry, properties of a molecule often depend on the interaction of the atoms at its oposite sides and their neighborhood topology~\cite{deepatoms}. 
These tasks require \emph{structural representations}, i.e. embeddings that can identify structural properties of a node's neighborhood. There is a growing literature that adresses this problem through different approaches. RolX ~\cite{rolx} extracts features for each node and performs non-negative matrix factorization to automatically discover node roles. Struc2vec~\cite{DBLP:journals/corr/FigueiredoRS17},
performs random walks on a constructed multi-layer graph to learn structural representations. GraphWave~\cite{donat} and DRNE~\cite{drne} employ diffusion wavelets and LSTM aggregation operators, respectively, to generate structural node embeddings.
However, most of these approaches suffer from high time or space complexity. 

In this paper, we propose a novel and simple structural node representation algorithm, \MN, that capitalizes on information-theoretic tools. The algorithm employs the Von Neumann entropy to construct node representations related to the structural identity of the neighborhood of each node. These representations capture the structural symmetries of the neighborhoods of increasing radius of each node. We show empirically the ability of \MN to identify structural roles and its robustness to graph perturbations through a node classification and node clustering study on highly symmetrical synthetic graphs.  
Moreover, we introduce a method of combining the generated representations by \MN with the node attributes of a graph, in order to avoid the incorporation of the graph topology in the optimization, contrary to the workflow of a GNN. Evaluated on real-world graph classification tasks, \MN achieves state-of-the-art performance, while maintaining a high efficiency compared to standard GNN models.

\begin{figure*}[ht]
    \centering
    \includegraphics[width=\textwidth,height=2.8cm]{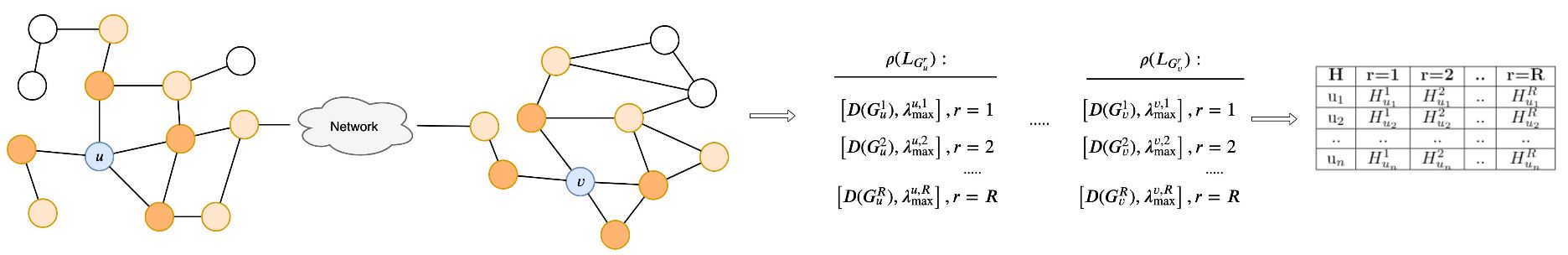}
    \caption{\MN extracts ego-networks for a series of defined radii and computes the VNE for every radius. On the left, the two parts of the network may have a large distance through the network. The 1-hop ego-networks are highlighted with the dark yellow, while the remaining nodes of the 2-hop ego-networks are highlighted with the light yellow. The nodes $u,v$ have structurally equivalent 1-hop neighborhoods $G_u^1$ and $G_v^1$, though their 2-hop neighborhoods $G_u^2$, $G_v^2$ do not.}
    \label{fig:method}
\end{figure*}

\section{Structural Representations based on Von Neumann Entropy}

Next, we present \MN for generating structural node representations, employing the Von Neumann entropy, a model-agnostic measure, that quantifies the structural complexity of a graph. The Von Neumann graph entropy (VNE) has been shown to have a linear correlation with other graph entropy measures~\cite{anand}. Graph entropy methods have been recently proved successful for computing graph similarity~\cite{lipan}.

\subsection{Von Neumann Entropy on Graphs}

In quantum mechanics, the state of a quantum mechanical system is described by a \emph{density} matrix $\rho$, i.e., a positive semidefinite, hermitian matrix with unit trace~\cite{Braunstein2006}. The Von Neumann entropy of the quantum system is defined as: 
\setlength\abovedisplayskip{1pt}
\setlength\belowdisplayskip{1.3pt}

\begin{equation}
    \label{eq::1}
    H(\rho) = -\Tr(\rho\log\rho) = -\sum_{i=1}^n \lambda_i\log\lambda_i,
\end{equation}
where $\Tr(\cdot)$ is the trace of a matrix, and $\lambda_i$'s are the eigenvalues of $\rho$.
Correspondingly, connecting it to graphs, given a graph $G=(V,E)$ and its Laplacian $L_G= D-A$, the VNE denoted by $H(G)$ , is defined as in Equation \ref{eq::1}, by replacing $\rho$ with $\rho(L_G) = \frac{L_G}{\text{Tr}(L_G)} = \frac{L_G}{2|E|}$~\cite{Braunstein2006}.
Note that $\lambda_i = \frac{1}{\text{Tr}(L_G)}v_i$ where $\lambda_i, v_i$ are the $i$-th eigenvalue of $\rho(L_G)$ and $L_G$, respectively. Therefore, $0 \leq \lambda_i \leq 1$ holds for all $i \in \{ 0,1,\ldots,n \}$~\cite{Passerini2009QuantifyingCI}.
This indicates that \Eqref{eq::1} is equivalent to the Shannon entropy of the probability distribution $\{\lambda_i\}_{i=1}^n$.
Hence, $H(G)$ serves as a skewness metric of the eigenvalue distribution and it has been shown that it provides information on the structural complexity of a graph~\cite{Passerini2009QuantifyingCI}.

\textbf{Efficient approximation scheme.}
The computation of VNE requires the eigenvalue decomposition of the density matrix which can be done in $\mathcal{O}(n^3)$ time. Recent works~\cite{fast_incr, choi2018fast} have proposed an efficient approximation of $H(G)$. Starting from \Eqref{eq::1} and following~\cite{minello}, we obtain:
\begin{equation}\label{eq::3}
    H(G) \approx \text{Tr}\big(\rho(L_G)(I_n - \rho(L_G))\big) = Q,
\end{equation}
where $I_n$ is the $n \times n$ identity matrix, and
\setlength\abovedisplayskip{1pt}
\setlength\belowdisplayskip{1pt}

\begin{equation}\label{eq::4}
        Q = \frac{\text{Tr}(L_G)}{2m} - \frac{\text{Tr}(L_G^2)}{4m^2}
        = 1 - \frac{1}{2m} - \frac{1}{4m^2}\sum_{i=1}^n d_i^2\,,
\end{equation}
where $m=|E|$ and $d_i$ is the ith-node degree. Finally, as~\cite{fast_incr} suggests, we obtain a tighter approximation of $H(G)$:
\setlength\abovedisplayskip{1.3pt}
\setlength\belowdisplayskip{1.3pt}

\begin{equation}\label{eq::5}
 \hat{H} = -Q\ln\lambda_{\max}\,,
\end{equation}

where $\lambda_{\max}$ is the largest eigenvalue of $\rho(L(G))$.
It can be shown that for any graph $G$, we have $H(G) \geq \hat{H}(G)$ where the equality holds if and only if $\lambda_{\max} = 1$~\cite{choi2018fast}.

\begin{table*}[t]
\centering
\def\arraystretch{1.1}
\resizebox{\textwidth}{!}{
\begin{tabular}{|lc|l||ccc|cc|} \hline
Configuration & Shapes & Algorithm & Homogeneity & Completeness & Silhouette & Accuracy & F$1$-score \\ \hline
\multirow{7}{*}{Basic / Basic Perturbed} & \multirow{7}{*}{\includegraphics[width=.2\textwidth]{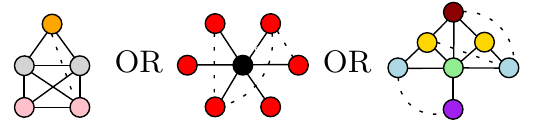}} & DeepWalk & 0.178 / 0.172 & 0.115 / 0.124 & 0.163 / 0/171 & 0.442 / 0.488 & 0.295 / 0.327 \\
& & RolX& 0.983 / 0.764 & 0.976 / 0.458 & 0.846 / 0.429 & \textbf{1.000} / 0.928 & \textbf{1.000} / \textbf{0.886} \\
& & struc2vec& 0.803 / 0.625 & 0.595 / 0.543 & 0.402 / 0.429 & 0.784 / 0.703 & 0.708 / 0.632 \\
& & GraphWave& 0.868 / 0.714 & 0.797 / 0.326 & 0.730 / 0.287 & 0.995 / 0.906 & 0.993 / 0.861 \\
& & VNEstruct& \textbf{0.986} / \textbf{0.882} & \textbf{0.983} / \textbf{0.701} & \textbf{0.891} / \textbf{0.478} & 0.920 / \textbf{0.940} & 0.901 / 0.881 \\ \hline

\multirow{7}{*}{Varied / Varied Perturbed} &
\multirow{7}{*}{\includegraphics[width=.2\textwidth]{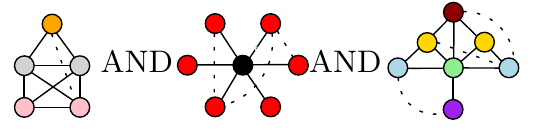}} & DeepWalk& 0.327 / 0.300 & 0.220 / 0.231 & 0.216 / 0.221 & 0.329 / 0.313 & 0.139 / 0.128 \\
& & RolX& \textbf{0.984} / 0.682  & 0.939 / 0.239 & 0.748 / 0.062 & \textbf{0.998} / 0.856 & 0.996 / 0.768 \\
& & struc2vec& 0.805 / 0.643 & 0.626 / 0.524 & 0.422 / \textbf{0.433} & 0.738 / 0.573 & 0.592 / 0.412 \\
& & GraphWave& 0.941 / 0.670 & 0.843 / 0.198 & \textbf{0.756} / 0.005 & 0.982 / 0.793 & \textbf{0.965} / 0.682 \\ 
& & VNEstruct& 0.950 / \textbf{0.722}  & \textbf{0.945} / \textbf{0.678} & 0.730 / 0.399 & 0.988 / \textbf{0.899} & 0.95 \textbf{0.878} \\ \hline 
\hline 
\end{tabular}
}
\caption{Performance of the baselines and the VNEstruct method for learning structural embeddings averaged over $20$ synthetically generated graphs. Dashed lines denote perturbed graphs.}
\label{tab:shapes}
\end{table*}

\subsection{The \MN Algorithm}
Based on the VNE and its approximation, we introduce our proposed approach to construct structural representations. The \MN algorithm extracts ego-networks of increasing radius and computes their VNE.
Then, the representation of a node comprises of the Von Neumann entropies that emerged from the node's ego-networks.
Therefore, the set of entropies of the ego-networks of a node serves as a ``signature'' of the structural identity of its neighborhood.

Let $R$ be the maximum considered radius. For each $r \in \{1,..,R\}$ and each node $v \in V$, the algorithm extracts the $r$-hop neighborhood  $G_v^r = (V',E')$, where $V' = \{u \in V | d(u,v) \leq r\}$ and $E' = \{(u,v) | u, v \in V', (u,v) \in E \}$.
Next, $H(G_v^r)$ of the $r$-hop neighborhood of $v$ is computed using \Eqref{eq::5}.
Finally, the $R$ entropies are arranged into a single vector $h_v \in \mathbb{R}^R$. As shown in Figure~\ref{fig:method}, \MN identifies structural equivalences of nodes that are distant to each other. Specifically, nodes $u$ and $v$ share structurally identical $1$-hop neighborhoods.
Therefore, the entropies of their $1$-hop neighborhoods are equal to each other.

\makeatletter
\newcommand{\removelatexerror}{\let\@latex@error\@gobble}
\makeatother

 
    

\textbf{Computational Complexity.}
The algorithm consists of: ($1$) the extraction of the ego-networks and ($2$) the computation of VNEs per subgraph.
The first step is linear in the number of edges of the node's neighborhood.
In the worst case, the complexity is $\mathcal{O}(nm)$, but for sparse graphs the complexity is constant in practice.
For the second step, following the approximation scheme in subsection 2.1, 
$\lambda_{max}$ is computed through the power iteration method~\cite{power_iteration}, which requires $\mathcal{O}(n+m)$ operations, as the Laplacian matrix has $n+m$ nonzero entries. 
Hence, the whole method exhibits linear complexity $\mathcal{O}(n+m)$, while for very sparse graphs, it becomes $\mathcal{O}(n)$.  

\textbf{Robustness over "small" perturbations.} We will next show that utilizing the VNE, we can acquire robust structural representations over possible perturbations on the graph structure. Clearly, if two graphs are isomorphic to each other, then their entropies will be equal to each other. It is important, though, for structurally similar graphs to have similar entropies, too. So, let $\rho, \rho' \in \mathbb{R}^{n \times n}$ be the density matrices of two graph laplacians $L_G,L_{G'}$, as described above.
Let also $\rho = P \rho' P^\top + \epsilon$ where $P$ is an $n \times n$ permutation matrix equal to $\argmin_{P} || \rho - P \rho' P^\top ||_F$ and $\epsilon$ is an $n \times n$ symmetric matrix.
If $G,G'$ are nearly-isomorphic, then the Frobenius norm of $\epsilon$ is small. By applying the Fannes-Audenaert inequality~\cite{Audenaert_2007}, we have that:
\begin{equation*}
 |H(\tilde{\rho}) - H(\rho)| \leq \frac{1}{2}T \ln(n-1) + S(T),   
\end{equation*} where $T = ||\tilde{\rho}-\rho||_1$ is the trace distance between $\rho,\tilde{\rho}$ and $S(T) = -T\log T- (1-T)\log(1-T)$.
However, $||\tilde{\rho}-\rho||_1 = \sum_i|\lambda_i^{\tilde{\rho}-\rho}| \leq n||\tilde{\rho} - \rho ||_{op}$, where $|| \cdot ||_{op}$ is the operator norm.
Therefore, $|H(\tilde{\rho})- H(\rho)| \leq \frac{n}{2}ln(n-1)||\epsilon||_{op} + S(T)$, leading thus to a size-dependent upper bound of the difference between the entropies of structurally similar graphs.

\subsection{Graph-level Representations}
Next, we propose how the structural representations generated by \MN can be combined with node attributes to perform graph classifications tasks.
The majority of the state-of-the-art methods learn node representations using message-passing schemes~\cite{hamilton, xu2018powerful}, where each node updates its representation according to its neighbors' representations, utilizing the graph structure information.

In this work, we do not use any message-passing scheme and we ignore the graph structure.
Instead, we augment the node attribute vectors of a graph with the structural representations generated by \MN. 
Thus, information about the graph structure is implictly incorporated into the augmented node attributes. Given a matrix of node attributes $X \in \mathbb{R}^{n\times d}$, the approach performs the following steps:

\begin{itemize}
    \itemsep0em 
    \item Computation of $H_v \in \mathbb{R}^{n\times R}$.
    \item Concatenation of node attribute vectors with structural node representations: $X' = [X || H] \in \mathbb{R}^{n\times (d+R)}$.
    \item Aggregation of node vectors $X'$ into:\\ $H_G = \psi( \sum_{v \in V_G} \phi(X'_v)) $, where $\phi$ and $\psi$ are MLPs.
\end{itemize}

This approach is on par with recent studies that propose to augment the node attributes with structural characteristics to avoid performing message-passing steps~\cite{DBLP:journals/corr/abs-1905-04579}. 
In comparison to a GNN, this procedure reduces the computational complexity of the training procedure since each graph is represented as a set of node representations.

\section{Experiments}
Next, we empirically show the robustness that \MN exhibits to graph perturbations in subsection 3.1 and we evaluate its graph classification performance in subsection 3.2.

\subsection{Structural Role Identification}

In order to evaluate the robustness of the structural representations generated by our method, we measure its performance on perturbed synthetic datasets, which were introduced in~\cite{donat, DBLP:journals/corr/FigueiredoRS17}.
We perform both classification and clustering tasks with the same experimental setup as in~\cite{donat}.\\
\textbf{Dataset setup.} The generated synthetic datasets are identical to those used in~\cite{donat}. They consist of basic symmetrical shapes, as shown in Table~\ref{tab:shapes}, that are regularly placed along a cycle of length $30$. The \textit{basic} setups use 10 instances of only one of the shapes of Table~\ref{tab:shapes}, while the \textit{varied} setups use 10 instances of each shape, randomly placed along the cycle. 
The perturbed instances are formed by randomly rewiring edges.
The colors in the shapes indicate the different classes.\\
\textbf{Evaluation.} For the classification task, we measure the \textit{accuracy} and the \textit{F1-score}. 
For the clustering task, we report the $3$ evaluation metrics, that were also calculated in~\cite{donat}: the \textit{Homogeneity} evaluates the conditional entropy of the structural roles in the produced clustering result, the  \textit{Completeness} evaluates how many nodes with equivalent structural roles are assigned to the same cluster and the \textit{Silhouette} measures the intra-cluster distance vs. the inter-cluster distance.

\begin{figure}[t]
    \centering
    \includegraphics[width=0.3\textwidth]{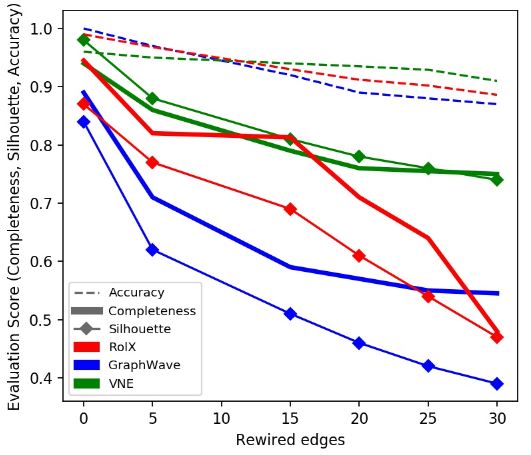}
    \caption{Classification and clustering performance of \MN and the baselines with respect to noise.}
    \label{fig:performances}
\end{figure}

In Table~\ref{tab:shapes}, VNEstruct outperforms the competitors on the perturbed instances of the synthetic graphs. On the basic and varied configurations, \MN outperforms the competitors in the node clustering evaluation and achieves comparable performance with RolX in node classification. On the perturbed configurations, \MN exhibits stronger performance than its competitors. The results in Table~\ref{tab:shapes} suggest a comparison of VNEstruct, RolX, and GraphWave in noisy scenarios. This comparison is provided in \Figref{fig:performances}, where we report the performance with respect to the number of rewired edges (from $0$ to $20$). We see that \MN is more robust than GraphWave and Rolx in the presence of noise.




\subsection{Graph Classification}
Next, we evaluate \MN and the baselines in the task of graph classification. We compare our proposed algorithm against well-established message-passing algorithms for learning graph representations.
Note that in contrast to most of the baselines, we pre-compute the entropy-based structural representations, and then we represent each graph as a set of vectors that encode structural characteristics. We used 4 common graph classification datasets (3 from bioinformatics: MUTAG, PROTEINS, PTC-MR and 1 from social-networks: IMDB-BINARY~\cite{xu2018powerful}). 

\textbf{Baselines.} The goal of the comparison is to show that by decomposing the graph structure and the attribute space, we can achieve comparable results to the state-of-the-art algorithms. Thus, we use as baselines graph neural network variants and specifically: DGCNN~\cite{Zhang2018AnED}, Capsule GNN~\cite{Xinyi2019CapsuleGN}, GIN~\cite{xu2018powerful}, GCN~\cite{Kipf:2016tc}, GAT~\cite{gat}. Moreover, GFN~\cite{DBLP:journals/corr/abs-1905-04579} augments the attributes with structural features and ignores the graph structure during the learning procedure.

\textbf{Model setup.}
For the baselines, we followed the same experimental setup, as described in~\cite{DBLP:journals/corr/abs-1905-04579} and, thus, we report the achieved accuracies. For GAT, we used a summation operator as an aggregator of the node vectors into a graph-level representation. Regarding the \MN, we performed 10-fold cross-validation with Adam optimizer and a 0.3 learning rate decay every 50 epochs. In all experiments, we set the number of epochs to 300. We choose the radius of the ego-networks from $r \in \{1,2,3,4\}$ and the number of hidden layers of the MLPs from $d \in \{8,16,32\}$.


Table~\ref{tab::graphclass} illustrates the average classification accuracies of the proposed approach and the baselines on the $5$ graph classification datasets.
Interestingly, the proposed approach achieves accuracies comparable to those of the state-of-the-art message-passing models. \MN outperformed all the baselines on $3$ out of $4$ datasets, while achieved the second-best accuracy on the remaining dataset (i.e., PROTEINS).

\begin{figure}[h!]
    \centering
    \includegraphics[width=1.\columnwidth, height=4.2cm]{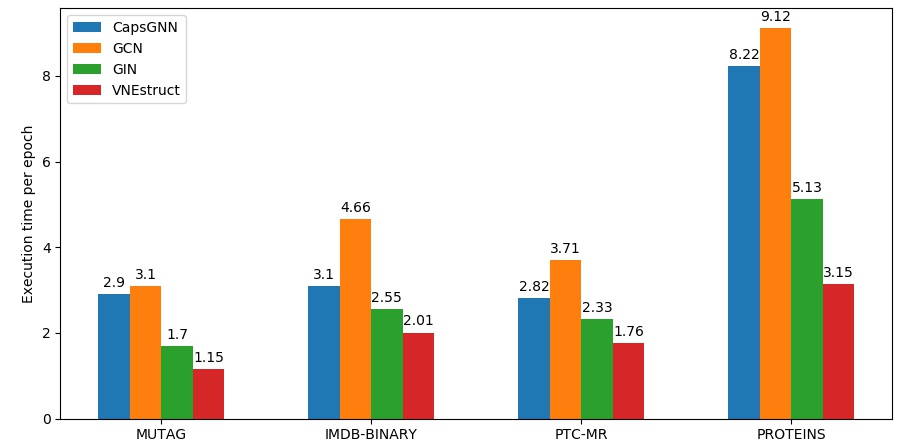}
    \caption{Training time per epoch (in sec) of VNEstruct and competitors for the graph classification tasks. }
    \label{fig:speed}
\end{figure}

\begin{table}[t]
\centering
\resizebox{0.47\textwidth}{!}{

\begin{tabular}{|c|cccc|}
\hline
  Method    & MUTAG   & IMDB-BINARY    & PTC-MR & PROTEINS     \\ \hline
  DGCNN      &  85.83 $\pm$ 1.66 & 70.03 $\pm$ 0.86 & 58.62$\pm$2.34 & 75.54 $\pm$ 0.94 \\ 
 
  CapsGNN      & 86.67 $\pm$ 6.88          & 73.10 $\pm$ 4.83         & - & 76.28 $\pm$ 3.63 \\ 
  GAT & 88.90 $\pm$ 3.21 & 75.39 $\pm$ 1.30 &  63.87 $\pm$ 5.31 &  76.1 $\pm$ 2.89 \\
  GIN & 89.40 $\pm$ 5.60 & 75.10 $\pm $ 5.10          & 64.6 $\pm$ 7.03 & 76.20 $\pm$ 2.60  \\ 
   GCN & 87.20 $\pm$ 5.11  & 73.30 $\pm$ 5.29 & 64.20 $\pm$ 4.30  & 75.65 $\pm$ 3.24 \\ 
\hline
  GFN & 90.84 $\pm$ 7.22 & 73.00 $\pm$ 4.29 & - &  \textbf{77.44 $\pm$ 3.77}\\ 
  \MN & \textbf{91.08 $\pm$ 5.65 }         & \textbf{75.40 $\pm$ 3.33}          & \textbf{65.39 $\pm$ 8.57}  & 77.41 $\pm$ 3.47  \\ 
  \hline
\end{tabular}
}
\caption{Average classification accuracy ($\pm$ standard deviation) of the baselines and the proposed \MN.}

\label{tab::graphclass}
\end{table}

Figure~\ref{fig:speed} illustrates the average training time per epoch of \MN and the baselines that apply message-passing schemes.
The proposed approach is generally more efficient than the baselines.
Specifically, it is $0.31$ times faster than GIN and $0.60$ times faster than GCN on average.
This improvement in efficiency is mainly because the graph structural features are computed in a preprocessing step and are then concatenated with the node attributes. However, the computational cost of the preprocessing step is negligible, as it is performed only once in the experimental setup. Furthermore, we should mention that due to the low dimensionality of the generated embeddings ($d \leq 4$), our method does not have any significant requirements in terms of memory.

\section{Conclusion}

In this paper, we proposed \MN to generate structural node representations, based on the entropies of ego-networks. We showed empirically the robustness of \MN under the presence of noise in the graph. We, also, proposed an approach for performing graph classification, that combines the representations of \MN with the nodes' attributes, avoiding the computational cost of message passing schemes. The proposed approach exhibited a strong performance in real-world datasets, maintaining high efficiency.

\vfill\pagebreak

\bibliographystyle{IEEEbib}
\bibliography{main_icassp}

\end{document}